\documentclass[11pt,a4paper]{article}
\usepackage[hyperref]{acl2019}
\usepackage{times}
\usepackage{latexsym}
\usepackage{url}

\usepackage{booktabs}
\usepackage{multirow}
\usepackage{rotating}
\usepackage{textcomp}
\usepackage[normalem]{ulem}
\usepackage{soul}
\usepackage{ragged2e} 

\usepackage{boxedminipage}

\usepackage{tikz}
\usepackage{pgfplots}

\newcommand{\normal}{\textsc{Standard}}
\newcommand{\reco}{\textsc{Suggestion}}

\newcommand{\cumulative}{\textsc{cum}}
\newcommand{\incremental}{\textsc{inc}}
\newcommand{\retrain}{\textsc{retrain}}

\mathchardef\mhyphen="2D

\aclfinalcopy 
\def\aclpaperid{729} 

\title{Analysis of Automatic Annotation Suggestions \\ for Hard Discourse-Level Tasks in Expert Domains}

\author{Claudia Schulz${}^1$, Christian M. Meyer${}^1$,
Jan Kiesewetter${}^2$,
Michael Sailer${}^3$, \\ \textbf{
Elisabeth Bauer${}^3$,
Martin R. Fischer${}^2$,
Frank Fischer${}^3$, and
Iryna Gurevych${}^1$} \\
${}^1$ Ubiquitous Knowledge Processing (UKP) Lab, Technische Universit\"at Darmstadt, Germany \\
${}^2$ Institute of Medical Education, University Hospital, LMU M\"unchen, Germany \\
${}^3$ Chair of Education and Educational Psychology, LMU M\"unchen, Germany \\
\url{http://famulus-project.de}
}

\date{}

\begin{document}
\maketitle
\begin{abstract}
Many complex discourse-level tasks can aid domain experts in their work but require costly expert annotations for data creation. 
To speed up and ease annotations, we investigate the viability of automatically generated annotation suggestions for such tasks.
As an example, we choose a task that is particularly hard for both humans and machines: the segmentation and classification of epistemic activities in diagnostic reasoning texts.
We create and publish a new dataset covering two domains and carefully analyse the suggested annotations. 
We find that suggestions have positive effects on annotation speed and performance, while not introducing noteworthy biases.
Envisioning suggestion models that improve with newly annotated texts, we contrast methods for continuous model adjustment and suggest the most effective setup for suggestions in future expert tasks.
\end{abstract}

\section{Introduction}
Current deep learning methods require large amounts of training data to achieve reasonable performance.
Scalable solutions to acquire labelled data use crowdsourcing (e.g., \citealp{Potthast18}), gamification (\citealp{Ahn06}), or incidental supervision \citep{Roth17}.
For many complex tasks in expert domains, such as law or medicine, this is, however, not an option since crowdworkers and gamers lack the necessary expertise.
Annotating data manually is therefore often the only way to train a model for tasks aiding experts with their work.
But the more expertise an annotation task requires, the more time- and funding-intensive it typically is, which is why many projects suffer from small corpora and deficient models.

In this paper, we propose and analyse an 
annotation setup aiming to increase the annotation speed and ease
for a discourse-level sequence labelling task requiring extensive domain expertise, without sacrificing annotation quality.
For the first time, we study the effects of automatically suggesting annotations to expert annotators in a task that is hard for both humans (only moderate agreement) and machine learning models (only mediocre performance) and compare the effects across different domains and suggestion models.
We furthermore investigate how the performance of the models changes if they continuously learn from expert annotations.

As our use case, we consider the task of annotating epistemic activities in diagnostic reasoning texts, which was recently introduced by \citet{SchulzEtAl2018_arxiv,SchulzMG2019}.
The task is theoretically grounded in the learning sciences \citep{FischerEtAl2014} and enables innovative applications that teach diagnostic skills to university students based on automatically generated feedback about their reasoning processes.
This task is an ideal choice for our investigations, since it is novel, with limited resources and experts available, and so far neural prediction models only achieve an F1 score of $0.6$, while also human agreement is in a mid range around $\alpha = 0.65$.

\citet{SchulzEtAl2018_arxiv} created annotated corpora of epistemic activities for 650 texts in the medicine domain (MeD) and 550 in the school teaching domain (TeD).
We extend these corpora by 457 and 394 texts, respectively.
As a novel component, half of the domain expert annotators receive automatically generated annotation suggestions.
That is, the annotation interface features texts with (suggested) annotations rather than raw texts. 
Annotators can accept or reject the suggested annotations as well as add new ones, as in the standard annotation setup.

Based on the collected data, we investigate the effects of these suggestions in terms of inter- and intra-annotator agreement, annotation time, suggestion usefulness, annotation bias, and the type of suggestion model.
As our analysis reveals positive effects,
we additionally investigate training suggestion models
that learn continuously as new data becomes available. Such incremental models can benefit tasks with no or little available data.

Our work is an important step towards our vision that even hard annotation tasks in expert domains, requiring extensive training and discourse-level context, can be annotated more efficiently, thus advancing applications that aid domain experts in their work.
Besides epistemic activities, discourse-level expert annotation tasks concern, for example, legal documents \citep{Nazarenko18}, psychiatric patient--therapist interactions \citep{Mieskes18}, or transcripts of police body cameras \citep{Voigt6521}.

The contributions of our work are:
(1) We study the effects of automatically suggesting annotations to expert annotators across two domains for a hard discourse-level sequence labelling task.
(2) We learn incremental suggestion models for little data scenarios through continuous adjustments of the suggestion model and discuss suitable setups.
(3) We publish new diagnostic reasoning corpora for 
two domains annotated with epistemic activities.\footnote{\url{https://tudatalib.ulb.tu-darmstadt.de/handle/tudatalib/2001}}

\section{Related Work}

\paragraph{Annotation Suggestions}
Previous work on automatic annotation suggestion (sometimes called pre-annotation) focused on token- or sentence-level annotations, including the annotation of part-of-speech tags \citep{FortS2010}, syntactic parse trees in historical texts \citep{EckhoffB2016}, and morphological analysis 
\citep{FeltEtAl2014}.
A notable speed-up of the annotation could be observed in these tasks, up to 70\,\% \citep{FeltEtAl2014}. However, \citet{FortS2010} find that annotation suggestions also biased the annotators' decisions.
\citet{Rosset2013} instead report no clear bias effects for their pre-annotation study of named entities. 
\citet{UlinskiHR2016} investigate the effects of 
different suggestion models for dependency parsing. 
They find that models with an accuracy of at least 55\,\% reduce annotation time.
Our work focuses on a different class of tasks, namely 
hard discourse-level tasks, in which the expert annotators only achieve a moderate agreement.

Annotation tasks in the medical domain are related to our use case in diagnostic reasoning.
\citet{LingrenEtAl2014} 
suggest medical entity annotations in clinical trial announcements.
\Citet{KholghiSZN2017} also investigate medical entity annotation, using active learning for their suggestion model, which results in a speed-up of annotation time.
\Citet{SouthEtAl2014} use automatic suggestions for de-identification of medical texts and find no change in inter-annotator agreement or annotation time.
In contrast to these works, 
we use a control group of two annotators,
who never receive suggestions, and compare the performance of all annotators to previous annotations they performed without annotation suggestions.

Work on the technical implementation of annotation suggestions is also still focused on
word- or sentence-level annotation types.
\Citet{MeursEtAl2011} use the GATE annotation framework \cite{BontchevaEtAl2013} for suggestions of biological entities.
\Citet{YimamBEG2014} describe the WebAnno system and discuss suggestions of part-of-speech tags and named entities using the MIRA algorithm \cite{CrammerS2003} for suggestion generation.
\Citet{SkeppstedtPK20176} introduce the PAL annotation tool, which provides suggestions and active learning for entities and chunks generated by logistic regression and SVMs. 
\Citet{GreinacherH2018} present the annotation platform DALPHI, suggesting named entity annotations based on a recurrent neural network.
Documents to be annotated are chosen by means of active learning,
enabling continuous updates of 
the
suggestion model during the annotation process. 
We also investigate continuous updates of the suggestion model during the annotation process, but focus on a task in which annotators require vast training and domain expertise.

\begin{figure*}[t]
  \newcommand{\MARK}[2]{\setlength{\fboxsep}{.75pt}\colorbox{#1}{\vphantom{Pg}#2}}
  \newcommand{\TYPEMARK}[1]{\textit{\uline{#1}}}
  \centering
  \begin{boxedminipage}{\textwidth}
    \raggedright
    \TYPEMARK{The patient reports to be lethargic and feverish}. \TYPEMARK{From the anamnesis I learned that he had purulent tonsilitis and is still suffering from symptoms}.
    \MARK{green}{I first performed some laboratory tests} and \TYPEMARK{notice the decreased number of lymphocytes}, \MARK{cyan}{which can be indicative of a bone marrow disease or an HIV} \MARK{cyan}{infection}.
    \TYPEMARK{The HIV test is positive}.
        \MARK{yellow}{However, the results from the blood cultures are negative,  so it is a} \MARK{yellow}{virus, parasite, or a fungal infection causing the symptoms}.
	\end{boxedminipage}
	\caption{Exemplary diagnostic reasoning text from the medicine domain, annotated with epistemic activity segments: \MARK{green}{evidence generation}, \TYPEMARK{evidence evaluation}, \MARK{yellow}{drawing conclusions}, \MARK{cyan}{hypothesis generation}.}
	\label{fig:reasoning_text}
\end{figure*}

\paragraph{Continuous Model Adjustment}
\citet{ReadBPH2012} distinguish two ways of training a model when new data becomes available continuously: 
using \emph{batches} or \emph{single} data points for the continuous adjustment, the latter often being referred to as online learning. We experiment with both adjustment strategies.
\citet{SanchesRB2010} propose \emph{incrementally} adjusting a neural network as new data becomes available, i.e.~only using the newly available data for the update. In addition to using incremental training, we also experiment with \emph{cumulative} training, where both previously available and new data is used for the model adjustment. 
\Citet{AndradeGR2017}, \citet{CastoEtAl2018}, and \citet{RusuEtAl2016} investigate adapting neural networks to new data with additional classes or even new tasks, requiring to change the structure of the neural network. Our setting is less complex as the neural network is trained on all possible classes from the beginning.
Recent work also investigates pre-training neural networks before training them on the actual data \cite{GargPJ2018,ShimizuSK2018,SerbanEtAl2016}. The model is thus adapted only once instead of continuously as in our work.

\section{Diagnostic Reasoning Task}
The annotation task proposed by \citet{SchulzEtAl2018_arxiv} has interesting properties for studying the effects of annotation suggestions in hard expert tasks: 
(1)~A small set of annotated data is available for two different domains.
(2)~Other than in well-understood low-level tasks, such as part-of-speech tagging or named entity recognition, the expert annotators require the discourse context to identify epistemic activities. This is a hard task yielding only inter-rater agreement scores in a mid range.
(3)~Prediction models only achieve F1 scores of around $0.6$, which makes it unclear if the suggestion quality is sufficient.

The previously annotated data consists of 650 German texts in the medical domain (MeD) and 550 texts in the teacher domain (TeD).
The texts were written by university students working on online case simulations, in which they had to diagnose the disease of a fictional 
patient (MeD),
or the cause of 
behavioural problems
of a fictional pupil (TeD) based on dialogues, observations, and test results.
For each case simulation, the students explained their diagnostic reasoning process in a brief self-explanation text.\footnote{Study approved by the university's ethics commission.}

Five (MeD) and four (TeD) domain experts annotated individual reasoning steps in the anonymised texts in terms of the epistemic activities
\cite{FischerEtAl2014}, 
i.e. activities involved in reasoning to develop a solution to a problem in a professional context.
We focus on the four most frequently used epistemic activities for our annotations:
hypothesis generation (HG), evidence generation (EG), evidence evaluation (EE), and drawing conclusions (DC).
HG is the derivation of possible diagnoses, which initiates the reasoning process.
EG constitutes explicit statements of obtaining evidence from information given in the case simulation or of recalling own knowledge. EE is the mentioning of evidence considered relevant for diagnosis.
Lastly, DC is defined as the derivation of a final diagnosis, which concludes the reasoning process.

As shown in Figure~\ref{fig:reasoning_text}, the annotation of these epistemic activities is framed as a joint discourse-level segmentation and classification task, that is, epistemic activities are segments of arbitrary length not bound to phrase or sentence level.

\section{Annotating with Suggestions}\label{sec:annotation_recommenders}

To conduct our experiments with annotation suggestions, we use the same annotation task and platform as \citet{SchulzEtAl2018_arxiv}.
We obtained their original data as well as further anonymised reasoning texts 
and we asked their expert annotators to participate in our annotation experiments. This allows us to study the effects of annotating data with and without suggestions, without having to account for changes in annotation performance due to individual expert annotators.

\begin{figure}[t]
    \centering
    \includegraphics[width=\linewidth,trim=0cm 1.75cm 7.2cm 0cm,clip]{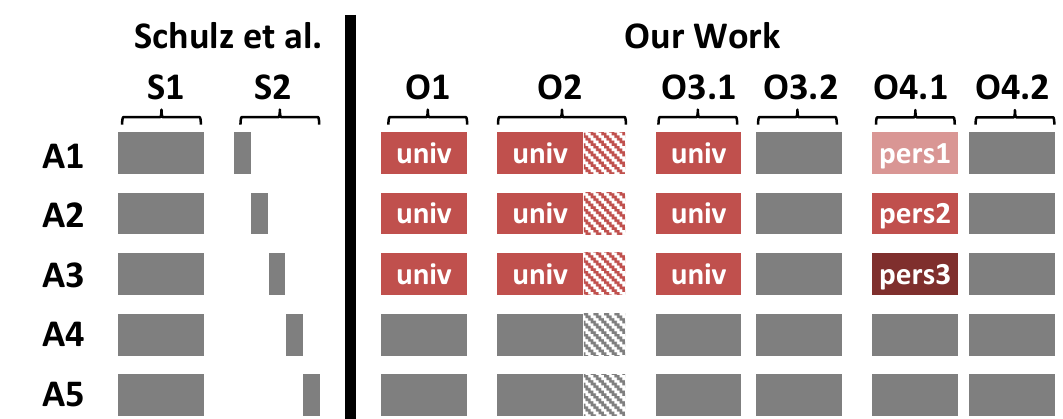}
    \caption{Annotation setup in MeD: Red indicates suggestions by a univ(ersal) or pers(onalised) model. The dashed boxes indicate annotations of texts that were already annotated in S1 or O1.}
    \label{fig:annotation}
\end{figure}

In total, we annotate 457 (MeD) and 394 (TeD) new reasoning texts during our experiments.
Figure~\ref{fig:annotation} shows an overview of our annotation phases in MeD with the five expert annotators A1 to A5.
S1 and S2 indicate the previous annotation phases by \citet{SchulzEtAl2018_arxiv}.
In their work, all experts first annotated the same 
texts (S1) and then
a different set of texts each (S2).
In our work, all experts annotate the same texts in all phases. We provide annotation suggestions to annotators A1 to A3 (randomly chosen among the five annotators) and instruct them to only accept epistemic activities if these coincide with what they would have annotated without suggestions and else manually annotate the correct text spans.
We study the effectiveness of the suggestions (O1), the intra-annotator consistency (O2), the annotation bias induced by suggestions (O3), and the effectiveness of a personalised suggestion model (O4).
Annotators A4 and A5 act as a control group, never receiving suggestions.
We use an analogous setup for TeD except that there is no annotator A3.

To create gold standard annotations, we use majority voting and annotator meetings as \citet{SchulzEtAl2018_arxiv}, and we publish our final corpora.

\subsection{Implementation}
\begin{figure}[ht]
     \centering
     \includegraphics[width=\linewidth,trim=1.1cm 0cm 5.3cm 1cm,clip]{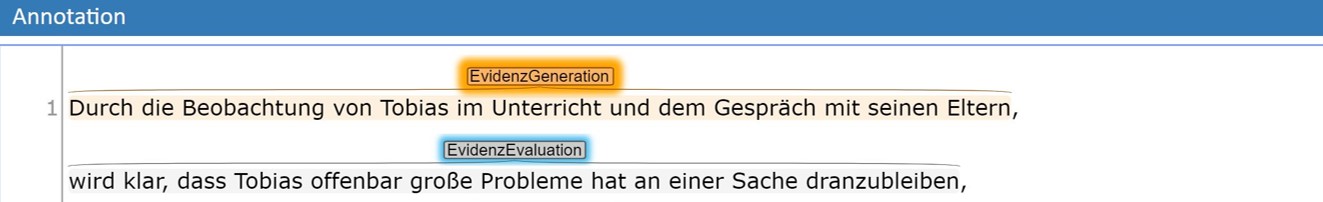}
     \caption{Annotation suggestion (grey) and accepted suggestion (orange) in the INCEpTION platform.}
     \label{fig:reco}
 \end{figure}

\paragraph{Annotation Tool}
Since we work with the same expert annotators as \citet{SchulzEtAl2018_arxiv},
we choose to also use the same annotation platform, \mbox{INCEpTION} \cite{KlieEtAl2018}, so that the expert annotators are already familiar with the interface.
 INCEpTION furthermore provides a rich API to integrate our suggestion models.
 As shown in Figure~\ref{fig:reco},
 annotation suggestions are shown in grey, distinguishing them clearly from differently coloured manual annotations. Suggestions can be easily accepted or rejected by single or double clicking. Additionally, manual annotations can be created as usual.

\paragraph{Suggestion Models}
\label{sec:recommender_model}
To suggest annotations, we use a state-of-the-art BiLSTM network with a conditional random field output layer \cite{ReimersG2017}, which has proven to be a suitable architecture for related tasks \cite{AjjourCKWS2017,EgerDG2017,LevyEtAl2018}.
We train this model using 
the gold standard of \citet{SchulzEtAl2018_arxiv}, consisting of annotations for all texts from phases S1 and S2.
The learning task is framed as standard sequence labelling with a BIO-encoding (Begin, Inside, Outside of a sequence) for the four epistemic activities hypothesis generation (HG), evidence generation (EG), evidence evaluation (EE), and drawing conclusions (DC).
More precisely, each token is assigned one of the labels $(\{B,I\} \times \{HG, EG, EE, DC\}) \cup \{O\}$, where $B \mhyphen HG$ denotes the first token of a HG segment, $I \mhyphen HG$ denotes a continuation token of a HG segment (similarly for $EG$, $EE$, and $DC$), and $O$ denotes a token that is not part of any epistemic activity.\footnote{We utilise the non-overlapping gold annotations of \newcite{SchulzEtAl2018_arxiv}, where a preference order over epistemic activities was applied to avoid overlapping segments.}
We use this suggestion model in O1--O3.1 and call it \emph{universal} (univ), as it learns labels obtained from all annotators of a domain.

For annotation phase O4.1, we train a \emph{personalised} (pers) suggestion model for each annotator A1--A3, based on the epistemic activities identified by the respective annotator in phases S1 and S2.
A personalised model thus provides suggestions tailored to a specific annotator. The idea of personalised models is that they may enable each annotator to accept more suggestions than possible with the universal model, which may lead to a speed-up in annotation time.
Note, however, that each of these personalised models is trained using only 250 texts, 150 annotated by the respective annotator in S1 and 100 in S2. Instead, the universal model is trained using 650 (MeD) or 550 (TeD) texts.

We train ten models with different seeds for each setup (universal and three personalised for MeD and TeD), applying the same parameters for all of them:
one hidden layer of 100 units, variational dropout rates for input and hidden layer of 0.25, and the \emph{nadam} optimiser \cite{nadam}.
We furthermore use the German \emph{fastText} word embeddings \cite{GraveBGJM2018} to represent the input. 
We apply early stopping after five epochs without improvement.
For the actual suggestions in our experiments, we choose the model with the best performance among the ten for each setup.

\subsection{Suggestion Quality}

Epistemic activity identification is a particularly hard discourse-level sequence labelling task, both for expert annotators and machine learning models.
Before beginning with our annotation experiments, we evaluate our different suggestion models, as shown in
Table~\ref{tab:f1_universal_annotator}.
All models exhibit mid-range prediction capabilities, which we consider sufficient for automatic annotation suggestions. 
This is supported by \citet{GreinacherH2018}, who find that suggestion models with an accuracy of at least 50\,\% improve annotation performance and speed for named entity recognition.
Still, the overall performance for our task is clearly lower than in low-level tasks such as part-of-speech tagging,
for which suggestions have been studied.

\begin{table}[ht]
\small
    \centering
    \begin{tabular}{ll cccc}
        \toprule
        Domain & Test Data & univ & pers1 & pers2 & pers3\\
        \midrule
        \multirow{2}{*}{MeD} & gold data & 0.63 &  0.51 & 0.58 & 0.55\\
        & ann. data & --- & 0.51 & 0.60 & 0.58\\
        \midrule
         \multirow{2}{*}{TeD} & gold data & 0.55 & 0.54 & 0.48 & ---\\
        & ann. data & --- & 0.60 & 0.49 & ---\\
        \bottomrule
    \end{tabular}
    \caption{Macro-F1 scores of the univ and pers models used in our experiments, evaluated on the gold and respective annotator-specific (ann.) annotations.}
    \label{tab:f1_universal_annotator}
\end{table}

We evaluate the performance of the personalised models using both the annotations by the respective annotator and the gold annotations.
The overall lower performance on the gold data shows that the personalised models indeed learn to predict the annotation style of the respective annotator.
We also observe lower performance of the personalised models compared to the universal models, which can be attributed to the smaller amount of annotated texts used for training.

\subsection{Evaluation and Findings}
 In this section, we examine the effects of annotation suggestions in detail, considering inter-annotator agreement, intra-annotator consistency, annotation bias and speed, as well as usefulness of suggestions and the impact of universal versus personalised suggestion models.

\paragraph{Effectiveness of Suggestions}
Since the annotation of epistemic activities involves determining spans as well as labels, we
measure the inter-annotator agreement (IAA) in terms of Krippendorff's $\alpha_{U}$ \cite{Krippendorff1995} as implemented in DKPro Agreement \cite{MeyerMSG2014}.
To evaluate the effects of suggestions on the annotations of our experts, we compare the IAA between annotators \emph{with} suggestions (A1--A3) -- henceforth called the \reco{} group -- against the IAA between annotators \emph{without} suggestions (A4--A5) -- denoted as the \normal{} group.
Table~\ref{tab:iaa} details the IAA of the two groups across all annotation phases described in Figure~\ref{fig:annotation}.

\begin{table}[ht]
\small
    \centering
    \begin{tabular}{l cccccc}
    \toprule
        & \multicolumn{3}{c}{MeD} &  \multicolumn{3}{c}{TeD}\\
        \cmidrule(lr){2-4} \cmidrule(lr){5-7}
        Phase & \textsc{St} & \textsc{Su} & \textsc{Su/St} & \textsc{St} &  \textsc{Su} & \textsc{Su/St}\\
         \midrule
         S1 & 0.65 & 0.67 & 0.67 & 0.65 & 0.64 & 0.65 \\ \midrule
         \textbf{O1} & 0.71 & 0.73 & 0.70 & 0.73 & 0.77 & 0.73\\
         \textbf{O2} & 0.66 & 0.69 & 0.64 & 0.66 & 0.76 & 0.67\\
         \textbf{O3.1} & 0.60 & 0.60 & 0.59 & 0.73 & 0.80 & 0.71\\
         O3.2 & 0.57 & 0.62 & 0.65 & 0.64 & 0.66 & 0.65 \\
         \textbf{O4.1} & -0.47 & 0.43 & 0.21 & 0.67 & 0.72 & 0.65\\
         O4.2 & 0.60 & 0.68 & 0.60 & 0.67 & 0.74 & 0.71\\
         \midrule
         O1--O4 & 0.48 & 0.64 & 0.59 & 0.69 & 0.75 & 0.68\\
         \bottomrule
    \end{tabular}
    \caption{Inter-annotator agreement in terms of Krippendorff's $\alpha_U$ for \textsc{St(andard)} and \textsc{Su(ggestion)} and their inter-group agreement (\textsc{Su/St}). Bold: Phases in which models were used for \textsc{Su}.} 
    \label{tab:iaa}
\end{table}

First, we compare the overall IAA of both groups for the previous annotation phase S1 by \citet{SchulzEtAl2018_arxiv} and all of our annotation phases O1--O4.2.
We observe for TeD that the IAA of the \reco{} group is consistently higher than of the \normal{} group, as soon as annotators receive suggestions (starting in O1). 
Since the IAAs of the two groups were similar in S1, when no suggestions were given, we deduce that suggestions cause less annotation discrepancies between annotators
in TeD.
Below, we will investigate if this also introduces an annotation bias.
For MeD, results are less clear, since the \reco{} group achieves only slightly higher IAA scores in most phases.
Notable is the extreme outlier of the \normal{} group in O4.1.
This is due to one annotator, whose EE (evidence evaluation) annotations deviated substantially from the other annotators. 
Considering the average IAA of our experiments without O4.1, we obtain very similar scores for the \normal\ (0.63) and \reco\ (0.66) group.
Thus, there is little difference to reference phase S1, where \reco{} already yielded a 0.02 higher IAA.
However, below we discuss the helpfulness and time saving of suggestions even in MeD.

\paragraph{Intra-Annotator Consistency}
In O2, we mixed 100 new texts with 50 texts the annotators saw previously during S1 or O1, but we did not inform the annotators about this setup.
Table~\ref{tab:intraAgreement} shows the annotation consistency of each annotator in terms of \emph{intra}-annotator agreement computed on those 50 double-annotated texts.
Even a single annotator shows annotation discrepancies instead of perfect consistency, evidencing the difficulty of annotating epistemic activities. Since the intra-annotator agreement for annotators with suggestions (A1--A3) is similar to that without (A4--A5), we conclude that suggestions do not considerably change annotators' annotation decisions.

\begin{table}[ht]
\small
\centering
 \begin{tabular}{c @{\qquad} ccc @{\qquad} cc @{\qquad} c}
  \toprule
  & \multicolumn{3}{@{}c@{\qquad}}{\reco} & \multicolumn{2}{@{}c@{\qquad}}{\normal} &\\
  & A1 & A2 & A3 & A4 & A5 & av.\\
  \midrule
  MeD & 0.74 & 0.76 & 0.79 & 0.78 &  0.80 & 0.77\\
  TeD & 0.77 & 0.64 & --- & 0.72 & 0.70 & 0.71\\
  \bottomrule
 \end{tabular}
 \caption{Intra-annotator agreement (in terms of Krippendorff's $\alpha_U$) on double-annotated texts.}
 \label{tab:intraAgreement}
\end{table}

\paragraph{Annotation Bias} 
The higher IAA in the \reco\ compared to the \normal\ group in TeD may indicate an annotation bias, i.e.~a tendency that the \reco\ group prefers the predicted labels over the actual epistemic activities. 
We test this unwanted effect by comparing the human--machine agreement between the experts' annotations and the models' predictions (in terms of Krippendorff's $\alpha_U$) for both annotators with and without suggestions.
Table~\ref{tab:annotation_prediciton} shows that, in both MeD and TeD, annotators who receive suggestions, i.e.~\reco\ in O1--O3.1 and in O4.1, consistently have a slightly higher agreement of about $0.1$
than annotators without suggestions in these phases. This indicates an annotation bias due to suggestions. 
In MeD, this bias is preserved even if annotators do not receive suggestions anymore (\reco\ in O3.2 and O4.2), whereas in TeD the bias fades.

To further examine the gravity of the annotation bias,
we compute
the inter-group agreement, i.e.\ the average pairwise IAA between annotators \emph{with} and \emph{without} suggestions, denoted \textsc{Su/St} in Table~\ref{tab:iaa}.
We find that 
this agreement is similar to the agreement
within the \normal\ group for both MeD and TeD.
In other words, an annotator with and an annotator without suggestions have the same level of agreement as two annotators without suggestions.

As a next step, we analyse the differences in the label distributions of the predictions and the \reco\ and \normal\ annotations.
In MeD, the \reco\ annotators use $EE$ (evidence evaluation) labels slightly more often, which can also be observed for the predictions.
In TeD, the \reco\ annotators use fewer $EE$ labels, but more $HG$ (hypothesis generation) labels than \normal\ annotators, which again matches the tendency of the predicted labels.
This effect is, however, very small, since all label distributions are close to each other.
The Jensen-Shannon divergence (JSD) between the label distributions of the two annotator groups is consistently below $0.02$ in all suggestion phases (O1--O3.1, O4.1) with an average JSD of $0.011$ (MeD) and $0.009$ (TeD).
There is almost no difference to the JSD of the remaining phases ($0.009$ for MeD, $0.010$ for TeD), indicating that the difference between the groups cannot be attributed to the suggestions.

We also compute the JSD of the \reco\ group and the predictions as well as the JSD of the \normal\ group and the predictions and find an average difference of the JSDs of $-0.009$ for MeD and $<0.001$ for TeD, which indicates a small bias towards the suggested labels for MeD, but no obvious bias for TeD.

\begin{table}[t]
    \small
    \centering
    \begin{tabular}{ll rrrrrr}
    \toprule
        & & \multicolumn{3}{c}{MeD} &  \multicolumn{3}{c}{TeD}\\
        \cmidrule(lr){3-5} \cmidrule(lr){6-8}
       &  &  \textsc{Su} & \textsc{St} & diff. &  \textsc{Su} & \textsc{St} & diff.\\ \midrule
        \parbox[t]{1mm}{\multirow{7}{*}{\rotatebox[origin=c]{90}{\textbf{univ}}}} & 
        S1 & 0.65 & 0.67 & --0.02     & 0.55 & 0.52 & +0.03\\
        &\textbf{O1} & 0.64 & 0.56 & +0.08    & 0.52 & 0.42 & +0.10\\
        &\textbf{O2} & 0.55 & 0.48 & +0.07    & 0.50 & 0.42 & +0.08 \\
        &\textbf{O3.1} & 0.69 & 0.55 & +0.14 & 0.54 & 0.40 & +0.14\\
        &O3.2 & 0.52 & 0.45 & +0.07  & 0.51 & 0.49 & +0.02\\
        &O4.1 & 0.46 & 0.33 & +0.13 & 0.47 & 0.39 & +0.08\\
        &O4.2 & 0.53 & 0.49 & +0.04  & 0.40 & 0.40 & +0.00\\
        \midrule
        \parbox[t]{1mm}{\multirow{2}{*}{\rotatebox[origin=c]{90}{\textbf{pers}}}} & 
         \textbf{O4.1} & 0.42 & 0.30 & +0.12 & 0.49 & 0.41 & +0.08\\
       & O4.2 & 0.41 & 0.45 & --0.04  & 0.34 & 0.32 & +0.02\\
        \bottomrule
    \end{tabular}
        \caption{Average $\alpha_U$ of annotators (in \textsc{Su(ggestion)} and \textsc{St(andard)}) with predictions of the univ and their pers model and diff(erence) between the groups. Bold: Phases in which models were used for \textsc{Su}.}
    \label{tab:annotation_prediciton}
\end{table}

\begin{figure}
    \centering
    \includegraphics[width=\linewidth,trim=2cm 24.5cm 6cm 2cm,clip]{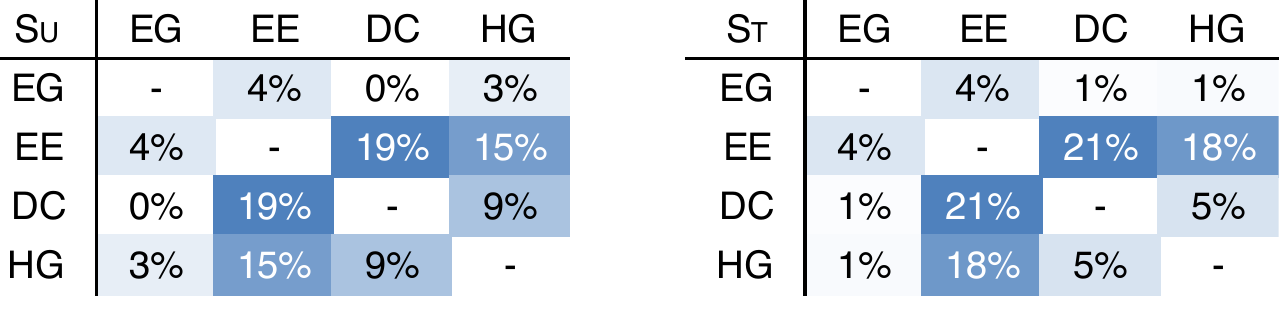}
    \caption{Disagreement among TeD annotators of the \textsc{Su(ggestion)} and \textsc{St(andard)} groups in phases with suggestions models (O1--O3.1 and O4.1).}
    \label{fig:disagreement}
\end{figure}

We finally analyse the disagreement within both groups of annotators.
Figure~\ref{fig:disagreement} shows the distribution of the disagreements for TeD's \reco\ (left) and \normal\ group (right).
We note that most disagreement occurs for $EE$ labels.
This is not surprising, as $EE$ is the most frequently occurring label.
The \reco\ group has a slightly higher disagreement for the $DC$ (drawing conclusions) and $HG$ labels, but overall, we do not observe substantial changes in the disagreement distribution, as also the disagreement for phases without suggestions is up to 3 percentage points different between the two groups.
For MeD, we find even smaller differences between the two groups.

Based on all analyses, we consider the annotation bias negligible, since suggestions do not cause negative annotation discrepancies compared to the standard annotation setup without suggestions.

\paragraph{Annotation Time}
 Table~\ref{tab:annotationTime} shows that nearly all annotators performed annotations faster in our experiments compared to previous annotations by \citet{SchulzEtAl2018_arxiv}, which can be attributed to the annotation experience they collected.
We note that annotators in the \reco\ group (A1--A3) always  speed up compared to previous annotations, whereas some of the annotators in the \normal\ group (A4--A5) slow down. Furthermore, on average, annotators in the \reco\ group exhibit a higher speed-up of annotation time: A1--A3 have a speed-up of 35\,\% compared to only 21\,\% for A4--A5 in MeD, and 20\,\% compared to only 11\,\% in TeD. Thus, suggestions make the annotation of epistemic activities more efficient.

\begin{table}[ht]
\small
\centering
 \begin{tabular}{ll ccc @{\qquad} cc}
 \toprule
  && \multicolumn{3}{c@{\qquad}}{\reco} & \multicolumn{2}{@{}c}{\normal}\\
 & Phase & A1 & A2 & A3 & A4 & A5 \\ 
 \midrule
   \parbox[t]{1mm}{\multirow{3}{*}{\rotatebox[origin=c]{90}{\textbf{MeD}}}} &
  S1--S2 & 1.92 &	2.13 &	1.82 &	3.78 &	1.94 \\
  & O1--O4 & 0.88 & 1.60 &	1.29 & 2.46 & 2.05 \\
  & speed-up & 54\,\% & 25\,\% & 29\,\% & 36\,\% & $-$6\,\% \\
 \midrule
  \parbox[t]{1mm}{\multirow{3}{*}{\rotatebox[origin=c]{90}{\textbf{TeD}}}} &
  S1--S2 & 2.73 & 2.91	& --- &	2.57 &	2.31\\
  & O1--O4 & 1.81 &  2.70 & --- & 2.76 & 1.59 \\
  & speed-up & 34\,\% &	7\,\% & --- &	$-$8\,\% &	31\,\%\\
\bottomrule
 \end{tabular}
 \caption{Average annotation time per text (in minutes) and speed-up of our compared to previous annotations.}
 \label{tab:annotationTime}
\end{table}

\paragraph{Usefulness of Suggestions}
In addition to positive informal feedback from the \reco\ annotators about the usefulness of suggestions, we also perform an objective evaluation.
As a new metric of usefulness, we propose the \emph{acceptance rate} of suggestions.
Table~\ref{tab:acceptedRecs} shows that on average 56\,\% of the suggestions are accepted by the expert annotators in MeD and 54\,\% in TeD.
Closer analysis reveals that in the many rejected cases, only the segment boundaries of suggestions were incorrect.
This leads us to conclude that suggestions ease the difficult task of annotating epistemic activities.

\begin{table}[ht]
    \small
    \centering
    \begin{tabular}{cccccc}
    \toprule
        &  O1 & O2 & O3.1 & O4.1 & av.\\
        \midrule
        MeD & 58\,\% & 49\,\% & 62\,\% & 54\,\% & 56\,\%\\
        TeD & 59\,\% & 55\,\% & 60\,\% & 43\,\% & 54\,\%\\
        \bottomrule
    \end{tabular}
    \caption{Percentage of accepted suggestions.}
    \label{tab:acceptedRecs}
\end{table}

\paragraph{Personalised versus Universal}
Both in MeD and TeD, Table~\ref{tab:iaa} shows a lower IAA in the \reco\ group when suggestions are given by a personalised model (O4.1) compared to the universal model (O1--O3.1). 
This can be explained by the fact that annotators are biased (see Table~\ref{tab:annotation_prediciton}, O4.1 pers) towards different annotations due to suggestions by different personalised models.

We observe that annotators also accept fewer suggestions from the personalised than from the universal models (see Table~\ref{tab:acceptedRecs}), which can be attributed to the worse prediction performance of the personalised models (see Table~\ref{tab:f1_universal_annotator}).
We conclude that our universal models exhibit more positive effects than the personalised models, as our goal is to create a gold standard corpus.

\paragraph{Discussion}
Our annotation study shows that annotation suggestions have various positive effects on the annotation of epistemic activities, despite the mediocre performance of our suggestion models.
In particular, the agreement between annotators in TeD is increased without inducing a noteworthy annotation bias, and
annotation time decreases in both MeD and TeD.
Since the task of epistemic activity identification is a particularly hard one, both for humans and for machine learning models, we expect that the positive effects of annotation suggestions generalise to other discourse-level sequence labelling tasks.

\section{Training Suggestion Models}
The previous section established that annotation suggestions have positive effects on annotating epistemic activities. However, these suggestions were only possible since \citet{SchulzEtAl2018_arxiv} had already annotated 550 reasoning texts in TeD and 650 in MeD, which were used to train our suggestion models.
Envisioning suggestions for similar tasks with fewer or even no existing annotations, this section simulates suggestions of our universal models in this scenario. We experiment with different methods of training our models with only a small number of `already annotated' texts and then continuously adjusting the models when `newly annotated' texts become available.

\subsection{Approach}
We use the gold annotations of \citet{SchulzEtAl2018_arxiv} for our experiments.
The ongoing annotation of texts and 
the continuously increasing amount of available training data can be simulated as a (random) sequence $S$ of texts $t_i$ becoming available at each time step $i$, i.e.~$S = t_1, t_2, \ldots, t_n$.

In addition to model adjustments at every time step, representing an online learning setup, we experiment with adjusting our models using \emph{bundles} of texts (called batches by \citet{ReadBPH2012}). The models are thus only adjusted after each $j$th time step, where $j$ is the bundle size.
We experiment with bundle sizes 10, 20, 30, 40, and 50 and represent the single-step setup as bundle size 1.

The easiest way to adjust a suggestion model for each new bundle is to train a new model from scratch using the union of the new and all previously available bundles.
We call this adjustment method \retrain\ and use bundle size 50. 
As a more advanced method, we suggest \emph{repeatedly training} the existing model every time a new bundle of texts becomes available, i.e.~the weights of the model are updated with each new bundle.
We contrast two strategies for updating the model:
the cumulative method (\cumulative) uses the union of the new and all previously available bundles of texts for training, whereas the incremental method (\incremental) uses only the new bundle.

For all model adjustment experiments, we use the architecture of our suggestion models described in Section~\ref{sec:recommender_model}. 
We report the average performance over ten runs for each setup (adjustment method, bundle size, domain).
Our text sequence $S$ has length 270. 
All models in the \cumulative\ and \incremental\ setup are initially trained on 10 texts before the repeated training with particular bundle sizes.

\begin{figure}[t]
    \centering
\resizebox {\columnwidth} {6.5cm} {
\begin{tikzpicture}
\begin{axis}[
    xlabel={{\small number of texts available}},
    xmin=0, xmax=280,
    ymin=0.25, ymax=0.6,
    tick label style={font=\small},
    legend pos=south east,
    ymajorgrids=true,
    grid style=dashed,
]
 
\addplot[
    very thick,
    color=red,
    ]
    coordinates 
    { (10, 0.279281743)	(20,0.318326993)	(30,0.41803026)	(40,0.460963864)	(50,0.478231467)	(60,0.483199459)	(70,0.50075135)	(80,0.513396714)	(90,0.518655307)	(100,0.521486321)	(110,0.533300392)	(120,0.545374612)	(130,0.552841427)	(140,0.551433187)	(150,0.56204033)	(160,0.560213978)	(170,0.566883865)	(180,0.578834109)	(190,0.576530797)	(200,0.570213109)	(210,0.578827514)	(220,0.581399531)	(230,0.586944466)	(240,0.5854691)	(250,0.589438423)	(260,0.586586196)	(270,0.59217078)
    };
    \addlegendentry{\small \cumulative}

 \addplot[
    thick,
    color=black,
    mark=square,
    ]
    coordinates 
   { (10,0.280934477)	(20,0.284195852)	(30,0.416012626)	(40,0.446976325)	(50,0.439256595)	(60,0.419054278)	(70,0.435557152)	(80,0.45486273)	(90,0.451737117)	(100,0.463660567)	(110,0.482436872)	(120,0.460041997)	(130,0.497723723)	(140,0.501158854)	(150,0.49379983)	(160,0.477666713)	(170,0.499912839)	(180,0.511256083)	(190, 0.492304379)	(200,0.476315681)	(210,0.48512755)	(220, 0.487108675)	(230,0.489713603)	(240,0.498817548)	(250,0.512508682)	(260,0.514616555)	(270,0.522103495)
    };
    \addlegendentry{\small \incremental\ 10}
    
     \addplot[
    thick,
    color=black,
    mark=triangle,
    ]
    coordinates 
   { (10,0.289826366) (40,0.469879644) (70,0.48271639) (100,0.498567345) (130,0.523984755) (160,0.515913059) (190,0.539900429) (220,0.535693752) (250,0.542751758)
    };
    \addlegendentry{\small \incremental\ 30}
    
     \addplot[
     thick,
    color=black,
    mark=o,
    ]
    coordinates 
   { (10,0.30089169) (60,0.48209811) (110,0.531818824) (160,0.548209787) (210,0.580222938) (270,0.592483066)
    };
    \addlegendentry{\small \incremental\ 50}
    
     \addplot[
     color=lightgray,
    ]
     coordinates
    {(10,0.3036649408)(11,0.2852392303)(12,0.1958200239)(13,0.2988687763)(14,0.3054050905)(15,0.2991126572)(16,0.2859656031)(17,0.248530417)(18,0.2673441209)(19,0.2264649076)(20,0.1984150105)(21,0.3132310734)(22,0.3374190891)(23,0.3084721262)(24,0.3393170737)(25,0.3458897827)(26,0.3195815665)(27,0.3210992718)(28,0.2768014487)(29,0.3118572031)(30,0.313278297)(31,0.3204859109)(32,0.3439025862)(33,0.3373114552)(34,0.3345923527)(35,0.3781630791)(36,0.3775403788)(37,0.3700395329)(38,0.3719559427)(39,0.3189828604)(40,0.3926288745)(41,0.3767609743)(42,0.3858781605)(43,0.3802383115)(44,0.3732390318)(45,0.3361997621)(46,0.3356678276)(47,0.3458388184)(48,0.3900303518)(49,0.3777528793)(50,0.3784210582)(51,0.3816129855)(52,0.357216028)(53,0.3444219263)(54,0.3745359017)(55,0.4113601096)(56,0.410280358)(57,0.4013969869)(58,0.3792887837)(59,0.3058862296)(60,0.3084406083)(61,0.2867378481)(62,0.3429161911)(63,0.4000459771)(64,0.3748751314)(65,0.3538160837)(66,0.4037362775)(67,0.3893699485)(68,0.4024812898)(69,0.375244449)(70,0.362644427)(71,0.3415753202)(72,0.3471779687)(73,0.3519858218)(74,0.3364961108)(75,0.3383346465)(76,0.4157492712)(77,0.4364113312)(78,0.4353866338)(79,0.4321436992)(80,0.4084832675)(81,0.4417220026)(82,0.4151337849)(83,0.4329254514)(84,0.3852463874)(85,0.4300642072)(86,0.3881039079)(87,0.3357924813)(88,0.3274077859)(89,0.3955602401)(90,0.4132698989)(91,0.4279824669)(92,0.4044822378)(93,0.4419677366)(94,0.396394062)(95,0.4252519285)(96,0.4115395964)(97,0.4183602277)(98,0.3909711434)(99,0.3878424105)(100,0.4094483238)(101,0.351405451)(102,0.3218774539)(103,0.3730213477)(104,0.3353554973)(105,0.4084949231)(106,0.4150821835)(107,0.441980888)(108,0.4325203639)(109,0.4260295728)(110,0.4308291216)(111,0.4144277982)(112,0.4053019973)(113,0.435915926)(114,0.4390067377)(115,0.4168799528)(116,0.4185946336)(117,0.4008836823)(118,0.4498310317)(119,0.4410456354)(120,0.4315097961)(121,0.4485771144)(122,0.4152098817)(123,0.4743496701)(124,0.4485506603)(125,0.4634052926)(126,0.4273154724)(127,0.4173429632)(128,0.4376167999)(129,0.419918965)(130,0.393977731)(131,0.3224919378)(132,0.3596579339)(133,0.3438202743)(134,0.4258082179)(135,0.4438792391)(136,0.4477303911)(137,0.4599798762)(138,0.4470176495)(139,0.4387828391)(140,0.4398484746)(141,0.4547609028)(142,0.446795485)(143,0.4472602537)(144,0.4651931151)(145,0.450910806)(146,0.4370398331)(147,0.4552405249)(148,0.4452987275)(149,0.4500922818)(150,0.449636741)(151,0.4115595245)(152,0.3853097381)(153,0.4168775786)(154,0.3951816462)(155,0.3475876415)(156,0.3614252791)(157,0.4156495145)(158,0.431962064)(159,0.4489202204)(160,0.4367250855)(161,0.4367291999)(162,0.4062206149)(163,0.4273547182)(164,0.3932203766)(165,0.3495815197)(166,0.4666651289)(167,0.4504052961)(168,0.4305065479)(169,0.4639680864)(170,0.4570907005)(171,0.4355562474)(172,0.466922274)(173,0.4543943647)(174,0.4695533059)(175,0.4607049974)(176,0.448978142)(177,0.4625192519)(178,0.4511074659)(179,0.4378608053)(180,0.4700018818)(181,0.4609450262)(182,0.4526550749)(183,0.4554036398)(184,0.4566856222)(185,0.4191749044)(186,0.4774040184)(187,0.4663746095)(188,0.4781827355)(189,0.46596942)(190,0.4153895924)(191,0.4213400854)(192,0.4522893963)(193,0.415570771)(194,0.4078743183)(195,0.4077109115)(196,0.4639226626)(197,0.4631423432)(198,0.4341999434)(199,0.4375006257)(200,0.4173249741)(201,0.4384378081)(202,0.4594364884)(203,0.4799356049)(204,0.460871148)(205,0.4582728748)(206,0.4325451414)(207,0.420161441)(208,0.4441291993)(209,0.4175728075)(210,0.4779539747)(211,0.4779278575)(212,0.4567736314)(213,0.3880747079)(214,0.4378555074)(215,0.4574323036)(216,0.4566299069)(217,0.4688319649)(218,0.4541752558)(219,0.4684877476)(220,0.4788085151)(221,0.4532147691)(222,0.4640300571)(223,0.4319357333)(224,0.4718119994)(225,0.4752149693)(226,0.473149038)(227,0.4485634973)(228,0.3931490516)(229,0.4484858911)(230,0.3748911277)(231,0.4354659671)(232,0.4520631072)(233,0.443147889)(234,0.4504435946)(235,0.4093640191)(236,0.4662087101)(237,0.4772753369)(238,0.4309905015)(239,0.4639800095)(240,0.4500724979)(241,0.4488539805)(242,0.4675994009)(243,0.4598461098)(244,0.4718273142)(245,0.4630156743)(246,0.4871726084)(247,0.4831711696)(248,0.4907251704)(249,0.4804224363)(250,0.473450176)(251,0.4716707096)(252,0.4631170925)(253,0.4836646045)(254,0.4544204857)(255,0.4710368203)(256,0.4886240551)(257,0.4890097826)(258,0.4868893558)(259,0.4906923964)(260,0.453425397)(261,0.4463574588)(262,0.4743949217)(263,0.4792904859)(264,0.4909834475)(265,0.5028800294)(266,0.5004191444)(267,0.4954242239)(268,0.4814179655)(269,0.4887680246)(270,0.4941498306)(271,0.4876486405)(272,0.501373243)(273,0.4975896057)(274,0.4788056461)(275,0.4823444509)
    };
    \addlegendentry{\small \incremental\ 1}
    
    \addplot[
    only marks,
     thick,
    color=blue,
    mark=asterisk,
    mark size = 5,
    ]
    coordinates 
   { (50,0.476719332) (100,0.523679973) (150,0.559603199) (200,0.573540174) (275,0.577269911)
    };
    \addlegendentry{\small \retrain}

\end{axis}
\end{tikzpicture}}
    \caption{Macro-F1 after each adjustment using different methods and bundle sizes in MeD.}
    \label{fig:retraining_f1}
\end{figure}
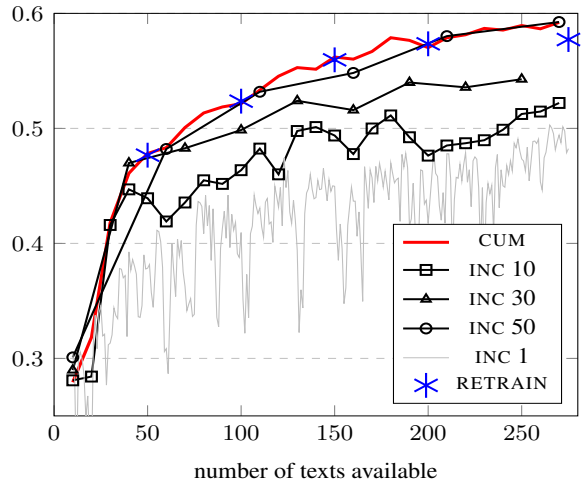

\subsection{Results and Evaluation}
We observe similar trends for MeD and TeD and therefore only present our MeD results in detail.

\paragraph{Model Performance} Figure~\ref{fig:retraining_f1} shows the macro-F1 scores for the different adjustment methods with various bundle sizes.
Using \cumulative, performance is very similar for all bundle sizes (1--50), thus represented by a single line in Figure~\ref{fig:retraining_f1}. \incremental\ with bundle sizes 20 and 40 are omitted from the figure for readability.
We observe that repeatedly training the model with \cumulative\ yields the same performance as \retrain, i.e.~as training a new model from scratch for every new bundle.
Furthermore, the performance of \cumulative\ rapidly increases with each bundle for the first 70 texts, reaching 0.5 macro-F1. The performance increase is more gradual thereafter, reaching 0.6 after 270 texts.

 Using \incremental\ for repeated training, bundle size influences performance: A small bundle size of 1 to 20 results in unsteady performance, which increases in the long-run but shows decreases after training on some of the bundles.
 In contrast, bundle sizes of 30 and higher show a steady increase in performance, similar to \cumulative. 
However, after having trained on at least 70 texts, \incremental\ adjustments with a bundle size smaller than 50 yield lower performance results than \cumulative\ adjustments.

We conclude that to provide annotation suggestions, repeatedly training a model using \incremental\ with a bundle size of 30 or more can be a suitable alternative to \cumulative\ as well as to training models from scratch whenever new annotations become available, since the performance sacrifice is small.

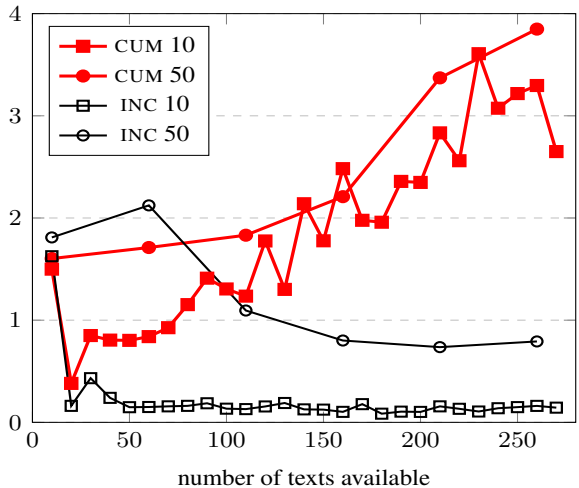
\begin{figure}[t]
    \centering
\resizebox {\columnwidth} {6.6cm} {
\begin{tikzpicture}
\begin{axis}[
    xlabel={{\small number of texts available}},
    xmin=0, xmax=280,
    ymin=0, ymax=240/60,
    tick label style={font=\small},
    legend pos=north west,
    ymajorgrids=true,
    grid style=dashed,
]
 
\addplot[
    very thick,
    color=red,
    mark=square*,
     y filter/.code={\pgfmathparse{\pgfmathresult/60}\pgfmathresult},
    ]
    coordinates 
    { (10,89.95551295) (20,22.87490597)	(30,50.9711056)	(40,48.296226)	(50,48.14726157)	(60,50.34279997)	(70,55.58532445)	(80,69.13128061)	(90,84.65878739)	(100,78.32788081)	(110,74.11297069)	(120,106.5128292)	(130,78.07648392)	(140,128.3800564)	(150,106.6150078)	(160,148.9777631)	(170,118.6460893)	(180,117.5200964)	(190,141.4918103)	(200,140.8967294)	(210,169.9914878)	(220,153.6281971)	(230,216.4807173)	(240,184.4572538)	(250,193.0651194)	(260,197.7956176)	(270,159.0000069)
    };
    \addlegendentry{\small \cumulative\ 10}

     \addplot[
    very thick,
    color=red,
    mark=*,
         y filter/.code={\pgfmathparse{\pgfmathresult/60}\pgfmathresult},
    ]
    coordinates 
    { (10,96.17440834) (60,102.6650788) (110,109.8926877) (160,132.5164628) (210,202.2805991) (260,230.9216781)
    };
    \addlegendentry{\small \cumulative\ 50}
 
 \addplot[
    thick,
    color=black,
    mark=square,
    y filter/.code={\pgfmathparse{\pgfmathresult/60}\pgfmathresult},
    ]
    coordinates 
  { (10,97.55579634) (20,9.686774731)	(30,25.9464783)	(40,14.38233488)	(50,8.903242469)	(60,9.011898756)	(70,9.401718974)	(80,9.689203906)	(90,11.20809603) (100,8.027613282)	(110,7.778902316)	(120,9.339977074)	(130,11.36702013)	(140,7.625168061)	(150,7.475158119)	(160,6.160069752)	(170,10.69311512)	(180,5.091695738)	(190,6.2902668)	(200,6.13286221)	(210,9.398731804)	(220,7.957094312)	(230,6.368470407)	(240,8.294111538)	(250,8.940059376)	(260,9.667199731)	(270,8.629311466)
    };
    \addlegendentry{\small \incremental\ 10}

     \addplot[
     thick,
    color=black,
    mark=o,
    y filter/.code={\pgfmathparse{\pgfmathresult/60}\pgfmathresult},
    ]
    coordinates 
  { (10,108.5924156) (60,127.4097423) (110,65.66568391) (160,48.05467184) (210,44.16711028) (260,47.51278834)
    };
    \addlegendentry{\small \incremental\ 50}
    
\end{axis}
\end{tikzpicture}}
    \caption{Training time (in minutes) for each adjustment using different methods and bundle sizes in MeD.}
    \label{fig:retraining_time}
\end{figure}

\paragraph{Training Time} 
Having observed only slight differences in the model performance using our different adjustment methods, Figure~\ref{fig:retraining_time} reveals a clear distinction regarding the time needed to adjust to each new bundle (trends of bundle sizes not illustrated lie between those in the figure).
While the training time using \cumulative\ \emph{increases} with each new bundle, since each successive adjustment is performed with more data, the training time of \incremental\ \emph{decreases} with each bundle, until reaching a stable minimum ranging from 8 seconds for bundle size 10 to 47 seconds for bundle size 50.
This decrease in training time, despite the stable amount of data used for each adjustment, is due to a decrease in the number of epochs required for training and indicates that the texts used in previous training steps are beneficial for training the model on a completely new bundle of texts.

The \retrain\ method, not illustrated in the figure, requires far more time for adjustment than the repeated training methods.
Training (from scratch) for 50 texts already takes 4.5 minutes, i.e.\ more than the \cumulative\ adjustment with 270 texts, and training for 270 texts takes 7.5 minutes.

\paragraph{Discussion} Our results show that \incremental\ adjustments are the most time-efficient, with each adjustment being two to five times faster than \cumulative\ adjustments.
In fact, in the \cumulative\ online learning setup (bundle size 1), the model adjustment time is similar to, and after 100 documents higher than the time needed for annotation (1--2 minutes per text as shown in Table~\ref{tab:annotationTime}).
However, the adjustment times of \cumulative\ with a bundle size of 10 or higher and of \retrain,
are lower than the time needed for annotating the respective bundle of texts.
Thus, \cumulative\ training with bundles larger than 1 is feasible for continuously adjusting suggestion models in our annotation task (while only a small amount of data is available), despite the long training time compared to \incremental. Since \cumulative\ achieves the same performance results as \retrain\ but needs far less time for adjustment, we dismiss \retrain\ as a suitable method for training suggestion models.

\section{Conclusion}
We presented the first study of annotation suggestions for discourse-level sequence labelling requiring expert annotators, using the hard task of epistemic activity identification as an example.
Our results show that even mediocre suggestion models have a positive effect in terms of agreement between annotators and annotation speed, while annotation biases are negligible.

Based on our experiments on training suggestion models,
we propose for future annotation studies that annotation suggestions can be given after having annotated only a small amount of data (in our case 70 texts), which ensures a sufficient model performance (0.5 macro-F1).
Since the exact number of texts required to reach sufficient model performance depends on the task, we suggest using continuous model adjustments from the start, ensuring flexibility as to when to start giving suggestions (namely whenever sufficient performance is achieved).
If computational resources are an important factor, we propose the usage of \incremental\ training with a bundle size of 30 or higher to optimise performance and training time. If model performance is more important, we recommend \cumulative\ training using a small bundle size of 10 or 20 to improve suggestions in short intervals.

In our model adjustment experiments, we used gold annotations.
To create them on the fly, annotation aggregation methods for sequence labelling \cite{SimpsonG2018} can be used.

We expect our work to have a large impact on future work requiring expert annotations, in particular regarding new tasks with no or little available data, for example for legal \cite{Nazarenko18}, chemical \cite{GuoEtAl2014}, or psychiatric \cite{Mieskes18} text processing.

\section*{Acknowledgements}
This work was supported by the German Federal Ministry
of  Education  and  Research  (BMBF)  under  the  reference
16DHL1040 (FAMULUS).
We 
thank our annotators
M.~Achtner, S.~Eichler, V.~Jung, H.~Mi\ss{}bach, K.~Nederstigt, P.~ Sch\"{a}ffner, R.~Sch\"{o}nberger, and H.~Werl.
We also acknowledge 
Samaun Ibna Faiz 
for his contributions to the model adjustment experiments.

\bibliography{acl2019_edas}
\bibliographystyle{acl_natbib}

\end{document}